\pdfoutput=1
\documentclass[sigconf,nonacm]{acmart}

\setcopyright{none}
\renewcommand\footnotetextcopyrightpermission[1]{}
\usepackage{booktabs}
\usepackage{tikz}
\usetikzlibrary{arrows.meta,positioning,fit,backgrounds,calc}
\usepackage{balance}

\begin{document}

\title{Overview of the EReL@MIR 2025 Multimodal Document Retrieval Challenge (Track 1)}

\author{Jingbiao Mei}
\affiliation{%
  \institution{University of Cambridge}
  \city{Cambridge}
  \country{United Kingdom}}
\email{jm2245@cam.ac.uk}

%% --- Concise CCS (edit as needed) ---


\begin{abstract}
Retrieval over visually-rich documents, pages that interleave text with
figures, tables, and charts, is essential for multimodal
retrieval-augmented generation, yet most retrievers still discard the
visual channel. The \emph{Multimodal Document Retrieval Challenge}, Track~1
of the MIR Challenge at the first EReL@MIR workshop, co-located with The Web
Conference 2025, asks participants to build a \emph{single} retrieval system
that handles two complementary regimes: closed-set document page retrieval within long
documents from a text query (MMDocIR), and open-domain retrieval of
Wikipedia-style passages from an image or image-plus-text query (M2KR).
Systems are ranked by the macro-average of mean Recall@$\{1,3,5\}$ over the
two tasks. The challenge drew 455 entrants and 586 submissions across 22 teams.
This report describes the challenge design, datasets, and evaluation protocol;
reports the final standings; and analyses the three winning teams' systems. All three
build on decoder-based Multimodal-LLM embedders from the Qwen2-VL family rather than on
CLIP-style encoders, and differ chiefly in whether they reach the top through
fine-tuned ensembles, training-free multi-route fusion with a strong
vision-language re-ranker, or zero-shot late interaction. The training-free
system finished within $0.1$ point of the fine-tuned winner.
\end{abstract}

\maketitle

%% =====================================================================
\section{Introduction}
%% =====================================================================
Real-world documents are multimodal: a single page of a financial
report, scientific paper, or encyclopedic web article interleaves prose
with tables, charts, figures, and layout structure. Answering questions over
such content with retrieval-augmented generation (RAG) requires a retriever
that can locate the correct \emph{page} or \emph{passage} from visual as well as
textual evidence. Text-only pipelines that rely on OCR discard much of this
signal, motivating a recent wave of vision-based retrievers that embed page
images directly~\cite{colpali,dse,gme}.

Progress, however, has been fragmented across two largely separate regimes.
The first is \emph{within-document} retrieval over long documents, where the
system must rank the relevant page(s) of one document given a text
query~\cite{mmdocir}. The second is \emph{open-domain} knowledge retrieval,
where a visual (or visual-plus-text) query must be matched against a large
global corpus of encyclopedic passages~\cite{preflmr}. The two regimes demand
different capabilities, fine-grained layout understanding versus
world-knowledge grounding, and are rarely evaluated under one model.

The EReL@MIR workshop (\emph{Efficient Representation Learning for
Multimodal Information Retrieval})~\cite{erelmir2025}, co-located with The Web
Conference 2025 in Sydney, organised the \emph{MIR Challenge} (MIRC) to bridge
this gap. Its Track~1, the Multimodal Document Retrieval Challenge, requires a
single unified model to serve \emph{both} regimes: Task~1 (MMDocIR) and Task~2
(M2KR). The unified-model constraint mirrors deployment reality---one
embedding system must index heterogeneous content---and the workshop's
emphasis on efficient, reusable representations.

This report provides the official overview of Track~1. It (i)~specifies the
two tasks, their datasets, and the evaluation protocol; (ii)~reports
participation statistics and the final leaderboard; (iii)~summarises the three
award-winning systems from their released code and reports;
and (iv)~distils the design choices that mattered. Our aim is to give a single
reference point for the benchmark and to draw out lessons for future editions.

%% =====================================================================
\section{Related Work}
%% =====================================================================
\textbf{Late interaction and visual document retrieval.}
ColBERT introduced \emph{late interaction}, scoring a query--document pair by
summing token-level maximum similarities (MaxSim) over multi-vector
representations~\cite{colbert,colbertv2}. ColPali and its ColQwen2 variant
carry this idea into the visual domain, encoding document-page \emph{images}
with a vision-language model and matching them with late interaction; ColPali
also introduces the ViDoRe benchmark~\cite{colpali}. Document Screenshot Embedding
(DSE) instead encodes each page screenshot into a single dense
vector~\cite{dse}, while VisRAG and M3DocRAG embed RAG pipelines around such
page-image retrievers~\cite{visrag,m3docrag}.

\textbf{Multimodal-LLM and fine-grained embedders.}
A parallel line repurposes multimodal LLMs as universal embedders: GME builds a
text/image/fused retriever on Qwen2-VL~\cite{gme,qwen2vl}, E5-V adapts MLLMs
with text-only training~\cite{e5v}, and VISTA augments a text encoder with
visual tokens~\cite{vista,bgem3}. FLMR and PreFLMR specialise late interaction
for knowledge-based visual question answering, projecting CLIP/SigLIP patch
features into a ColBERT token space~\cite{flmr,preflmr,clip,siglip}; PreFLMR is
the official baseline for the M2KR task (Task~2) and the origin of the M2KR
benchmark.
The challenge stress-tests these methods under a single
unified-model constraint, across two regimes that no prior benchmark
evaluated jointly.

%% =====================================================================
\section{Challenge Design}
%% =====================================================================

\subsection{Tasks}
\begin{figure}[t]
\centering
\begin{tikzpicture}[
  font=\footnotesize,
  box/.style={draw, rounded corners, align=center, inner sep=3pt, minimum height=6mm},
  qbox/.style={box, fill=blue!8},
  cbox/.style={box, fill=gray!10},
  model/.style={draw, rounded corners, fill=orange!12, align=center, minimum width=18mm, minimum height=12mm},
  arr/.style={-{Latex[length=2mm]}, thick},
]
% Unified model in the middle
\node[model] (m) {Unified\\multimodal\\retriever};

% Task 1 (top): text query -> pages within a doc
\node[qbox, above left=5mm and 6mm of m] (q1) {Text query};
\node[cbox, above right=5mm and 6mm of m] (c1) {Pages of\\one document};
\draw[arr] (q1) -- (m);
\draw[arr] (m) -- (c1);
\node[align=center] at ($(q1)!0.5!(c1)+(0,7mm)$) {\textbf{Task 1: MMDocIR} (within-document)};

% Task 2 (bottom): image / image+text -> global passage corpus
\node[qbox, below left=5mm and 6mm of m] (q2) {Image\\(+ text)};
\node[cbox, below right=5mm and 6mm of m] (c2) {Global passage\\corpus (47K)};
\draw[arr] (q2) -- (m);
\draw[arr] (m) -- (c2);
\node[align=center] at ($(q2)!0.5!(c2)-(0,7mm)$) {\textbf{Task 2: M2KR} (open-domain)};
\end{tikzpicture}
\caption{One unified model must serve two complementary retrieval regimes:
text-to-page retrieval within a long document (Task~1) and open-domain
image / image+text-to-passage retrieval over a global corpus (Task~2).}
\Description{A schematic with a central ``unified multimodal retriever'' box.
Above it, Task~1 (MMDocIR): a text query and the pages of a single document feed
into the model for within-document page retrieval. Below it, Task~2 (M2KR): an
image or image-plus-text query and a global 47K-passage corpus feed into the
model for open-domain retrieval.}
\label{fig:overview}
\end{figure}
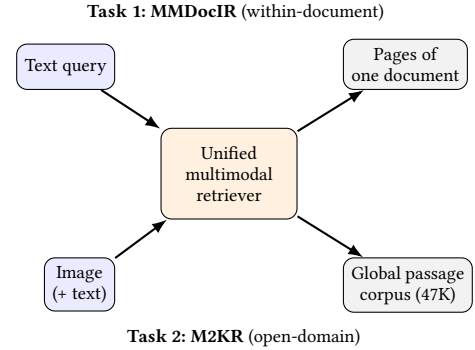

Track~1 couples two retrieval tasks that a participant must solve with one
system (Figure~\ref{fig:overview}).

\textbf{Task~1 -- MMDocIR (within-document page retrieval).}
Given a text-only query and a single long document, the system ranks the most
relevant \emph{pages} of that document~\cite{mmdocir}. Retrieval is scoped to
one document at a time---on average $65.1$ pages---rather than to the global
corpus, and the challenge uses the page-level (not layout-level) variant of
MMDocIR. Each page is provided as a screenshot together with its OCR text and a
VLM-generated description, so participants may retrieve visually, textually, or
both.

\textbf{Task~2 -- M2KR (open-domain visual retrieval).}
Given a query that is either an image alone or an image paired with a text
question, the system retrieves the relevant passage from a single global corpus
of $47{,}318$ encyclopedic, Wikipedia-style passages. Roughly half of the
queries carry a text question (image-plus-text-to-passage); the remainder are
purely visual (image-to-passage). Each corpus item couples a textual passage
with a web-page screenshot. M2KR draws on the WIT, OVEN, Infoseek, Encyclopedic-VQA,
and OK-VQA lineages~\cite{wit,oven,infoseek,evqa,okvqa,preflmr}.

\subsection{Datasets}
Table~\ref{tab:data} summarises the two challenge sets. During the challenge,
both withheld the ground-truth relevance labels of the test queries, which were
scored server-side on Kaggle; the M2KR ground truth has since been released.
M2KR-Challenge\footnote{\url{https://huggingface.co/datasets/Jingbiao/M2KR-WWW2025-Challenge}}
is a challenge-specific subset of the larger M2KR benchmark introduced with
PreFLMR~\cite{preflmr}. MMDocIR-Challenge\footnote{\url{https://huggingface.co/datasets/MMDocIR/MMDocIR-Challenge}}
provides 313 expert-curated long documents spanning ten domains (academic
papers, financial reports, government and legal documents, research reports,
guidebooks, tutorials, brochures, administrative/industry material, and
news)~\cite{mmdocir}.

\begin{table}[t]
\centering
\caption{The two Track~1 datasets. Test relevance labels were withheld during
the challenge.}
\label{tab:data}
\small
\begin{tabular}{@{}lcc@{}}
\toprule
 & \textbf{M2KR-Challenge} & \textbf{MMDocIR-Challenge} \\
 & (Task~2) & (Task~1) \\
\midrule
Regime            & open-domain         & within-document \\
Query modality    & image / image+text  & text \\
\# Test queries   & $6{,}415$           & $1{,}658$ \\
Corpus scope      & global              & per document \\
\# Passages/pages  & $47{,}318$          & $20{,}395$ \\
\# Documents       & ---                 & $313$ \\
Avg.\ pages/doc    & ---                 & $65.1$ \\
Corpus item       & passage + screenshot & page screenshot \\
                  &                      & \;+ OCR + VLM text \\
\bottomrule
\end{tabular}
\end{table}

\subsection{Evaluation Protocol}
For each task the score is the mean of Recall@$1$, Recall@$3$, and Recall@$5$;
recall is computed against the full set of relevant items, so a query with
multiple positives requires all of them in the top-$k$ to reach $100\%$.
The final leaderboard score is the macro-average of the two per-task scores,
weighting MMDocIR and M2KR equally regardless of their differing query counts.
Participants submit a ranked list of the top-5 passage (or page) identifiers
per query.\footnote{The official challenge scored the top-5 submissions, although
the MMDocIR dataset card describes a top-10 submission format.} Submissions were
evaluated on
Kaggle,\footnote{\url{https://www.kaggle.com/competitions/multimodal-document-retrieval-challenge}}
with a limit of three submissions per team per day; the public and private
leaderboards cover the same rows. As a condition of award, winning teams had to
submit their code for reproducibility and open-source it under the MIT or
Apache-2.0 license.

\subsection{Baseline}
PreFLMR~\cite{preflmr}, a fine-grained late-interaction retriever that projects
ViT-G/14 patch features into a ColBERT token space, is the official baseline
and the model from which M2KR originates. On the M2KR-Challenge test queries,
under the challenge's strict multi-positive scoring, PreFLMR\,(ViT-G) attains
Recall@$1/3/5$ of $19.5/34.0/41.2$, i.e.\ a mean of $31.6$. This open-domain
baseline anchors the harder of the two tasks; all three winning teams
substantially exceed it on the combined metric. No official baseline was
provided for the within-document MMDocIR task.

%% =====================================================================
\section{Participation and Results}
%% =====================================================================
The challenge attracted \textbf{455 entrants}, \textbf{41 active
participants}, \textbf{22 teams}, and \textbf{586 submissions}.\footnote{Participation
statistics and leaderboard scores are from the Kaggle competition dashboard,
accessed May 2025.} Table~\ref{tab:lb}
reports the top of the final leaderboard. The contest was closely fought at the
top: the first two teams finished within $0.1$ point of each other on the
combined score. The team ranked third on the leaderboard, GoAhead, was not
eligible for an award: the organisers could not verify its result without the
submitted code, as required for award eligibility. Third place therefore passed to
the next eligible team, ``GPU is all you need.''

\begin{table}[t]
\centering
\caption{Final leaderboard (top entries). Score is the combined macro-average of
mean Recall@$\{1,3,5\}$ over the two tasks ($\times 100$). Award rank reflects
the official results page.}
\label{tab:lb}
\small
\begin{tabular}{@{}clcc@{}}
\toprule
\textbf{Award} & \textbf{Team} & \textbf{Score} & \textbf{\#\,Sub.} \\
\midrule
1st & iLearn                       & $65.69$ & $120$ \\
2nd & LLMHunter                    & $65.59$ & $59$  \\
$\dagger$ & GoAhead                & $60.73$ & $62$  \\
3rd & GPU is all you need          & $57.30$ & $14$  \\
    & Ruifeng Hu                   & $50.90$ & ---   \\
    & hhudan                       & $49.01$ & ---   \\
\bottomrule
\end{tabular}
\\[2pt]
{\footnotesize $^{\dagger}$Ranked third on the leaderboard but ineligible for an
award (see text).}
\end{table}

%% =====================================================================
\section{Winning Solutions}
%% =====================================================================
We summarise each awarded system from its public code and report. All three
favour a multimodal-LLM embedder over CLIP-style or PreFLMR-style encoders,
motivated by the world knowledge that open-domain M2KR demands. The teams
differ mainly in how much they train and how they re-rank.

\subsection{First place: iLearn}
iLearn~\cite{ilearn} builds on \emph{GME-Qwen2-VL-7B}~\cite{gme}, a multimodal
embedder, and reaches the top through a fine-tuned ensemble augmented with a
visual matching step. Three ideas stand out. (1)~\emph{$k\times$[EOS]
pooling}: instead of pooling a single end-of-sequence token, the system appends
several EOS tokens and mean-pools their final hidden states, which the team
reports adds $2$--$3$ points to top-$k$ recall at negligible cost. (2)~A
\emph{five-expert ensemble}: five GME models are LoRA-fine-tuned with different
learning rates and epochs on a multi-positive contrastive objective, then
combined by rank fusion across both tasks---these five together constitute the
required ``unified'' model. (3)~A \emph{Visual Anchor Point} mechanism for
M2KR: each candidate page screenshot is segmented into sub-images, a DINOv2
encoder~\cite{dinov2} computes the maximum cosine similarity between the query
image and these sub-images, and a candidate whose similarity exceeds a
per-subset threshold is force-promoted to rank~1. This exploits the fact that
many M2KR query images recur, near-duplicated, inside the corpus screenshots.
MMDocIR is handled by the same GME experts without the DINOv2 step, ranking
pages within each document by fused-embedding similarity.

\subsection{Second place: LLMHunter}
LLMHunter~\cite{llmhunter} reaches essentially the same score with a
\emph{training-free} multi-granularity framework: every component is
off-the-shelf, and the gains come from multi-route recall and re-ranking rather
than fine-tuning. The backbone is again GME-Qwen2-VL-7B, complemented by
\emph{ColQwen2-7B} late interaction~\cite{colpali} for MMDocIR pages and a
\emph{Qwen2.5-VL-72B} re-ranker that judges candidate relevance with a binary
yes/no prompt. For M2KR, the system merges a GME image-plus-text-to-text route
with a layout-level image-to-image route, the latter cross-checked by classical
computer-vision verification (homography inlier ratios over SIFT/ORB matches);
the merged top candidates are then re-ranked by the 72B vision-language model.
For MMDocIR, three routes---ColQwen2 late interaction, GME layout-level
image retrieval with query-adaptive instructions, and GME text-to-text over OCR
chunks---are fused and re-ranked. The authors report an ablation ladder from a
GME-only baseline of $53.96$ to a final
$65.56$:\footnote{LLMHunter's self-reported $65.56$ differs marginally from its
$65.59$ leaderboard entry in Table~\ref{tab:lb}, likely a public/private split
or rounding.} vision-language re-ranking alone accounts for the last step
($64.1\!\to\!65.6$), and the full gain of roughly $12$ points over the
single-model baseline comes from multi-route fusion and re-ranking rather than
from any single model. Reaching this score with no training, LLMHunter
finished in a near-tie with the fine-tuned first-place system.

\subsection{Third place: GPU is all you need}
The third system~\cite{gputeam} is entirely \emph{zero-shot}: no fine-tuning,
no re-ranker, and no ensemble, using only the ColQwen2 (ColPali-family) late
interaction model~\cite{colpali} for both tasks. For MMDocIR, each page is
encoded both as an image and from its provided VLM-generated text; the two
multi-vector representations are concatenated along the token dimension into one
fused page embedding, and pages are ranked by late-interaction MaxSim against
the text query. For M2KR the team takes a purely \emph{image-to-image} route:
candidate images are obtained by scraping or cropping figures from the
Wikipedia corpus screenshots, ColQwen2's multi-vector outputs are mean-pooled
into single dense vectors, and an exact FAISS $L_2$ index returns nearest
neighbours that are mapped back to source articles. A carefully applied
off-the-shelf model thus sets a strong zero-shot baseline, though a clear gap
to the trained systems remains.

%% =====================================================================
\section{Discussion and Lessons Learned}
%% =====================================================================
\textbf{One backbone, two specialised pipelines.}
Every winning team satisfied the unified-model constraint by sharing a single
embedding backbone while specialising the retrieval pipeline per task---visual
anchoring and image-to-image matching for open-domain M2KR, late interaction
over page images for within-document MMDocIR. No top team relied on a single
undifferentiated pipeline for both tasks, suggesting that task-aware routing
over a shared backbone is more effective.

\textbf{MLLM embedders displaced CLIP and PreFLMR.}
All three winners chose Qwen2-VL-derived embedders (GME, ColQwen2) over the
CLIP-based PreFLMR baseline, citing the world knowledge that open-domain
retrieval demands. The baseline's late-interaction design nonetheless lived on,
re-expressed through ColQwen2.

\textbf{Two routes to the top, and an efficiency message.}
A fine-tuned five-expert ensemble (iLearn) and a training-free multi-route
framework with a strong re-ranker (LLMHunter) finished within $0.1$ point of
each other. The latter reaches the top of the leaderboard without updating a
single weight---a direct illustration of the workshop's premise that efficient
use of existing representations can match fine-tuning.

\textbf{Visual near-duplication is exploitable, and a caveat.}
The strongest M2KR gains came from matching query images to near-duplicate
regions inside corpus screenshots. This reflects a real property of
encyclopedic retrieval but also raises a dataset-construction concern for future
editions: corpora should be curated so that retrieval rewards semantic grounding
rather than pixel-level duplication.

\textbf{Limitations of the protocol.}
A single combined score is easy to communicate but hides per-task behaviour, and
the equal weighting of two tasks with very different query counts and
difficulties is a deliberate but debatable choice. Because the public team
reports provide only combined scores, the per-task split is not recoverable here;
requiring per-task Recall@$\{1,3,5\}$ reporting, and adding a layout-level track,
would sharpen future analysis.

%% =====================================================================
\section{Conclusion}
%% =====================================================================
The Multimodal Document Retrieval Challenge at EReL@MIR 2025 established a
unified benchmark that pairs within-document page retrieval with open-domain
visual retrieval and requires a single system for both. Across 22 teams, the
winning solutions converged on multimodal-LLM embedders and task-aware
pipelines, and showed that a training-free, re-ranking-centric system can match
a fine-tuned ensemble. We hope the released datasets, baselines, and open-sourced
winning systems serve as a reference for efficient multimodal retrieval, and we
look forward to future editions that report finer-grained metrics and broaden
the document domains.

%% ------------------------------------------------------------------
\begin{acks}
We thank the MIR Challenge organising team---Xuri Ge, Junchen Fu, Fajie Yuan,
Jieming Zhu, Jiabing Yan, Kuicai Dong, Zhaocheng Du, and Yong Liu---for
designing and running the challenge; the EReL@MIR~2025 workshop organising
committee---Xin Xin, Haitao Yu, Yue Feng, Alexandros Karatzoglou, Ioannis
Arapakis, and Joemon M.~Jose---and program committee---Hui Li, Qian Li, Siwei
Liu, Songpei Xu, Xi Wang, Jiayi Ji, Hengchang Hu, Fuhai Chen, and Mingyue
Cheng---for hosting it; and the participating teams and the contributors of the
M2KR and MMDocIR benchmarks. We gratefully acknowledge Huawei for providing
funding.
\end{acks}

\balance
\bibliographystyle{ACM-Reference-Format}
\bibliography{references}

\end{document}